\documentclass[11pt,a4paper]{article}
\usepackage{acl2015}
\usepackage{times}
\usepackage{latexsym}
\usepackage{url}
\usepackage{color}
\usepackage{multirow}
\usepackage{comment}
\usepackage{amsmath} 
\usepackage{amssymb}

\usepackage{etoolbox}
\usepackage{graphicx}
 
 \usepackage{enumitem}
\newenvironment{tightitemize}%
  {\begin{itemize}[topsep=0pt, partopsep=0pt] %
    \setlength{\itemsep}{0pt}%
    \setlength{\parskip}{0pt}%
    }%
  {\end{itemize}}
  
  \usepackage{color}
  {\begin{enumerate}≈ \footnotesize%
    \setlength{\itemsep}{0pt}%
    \setlength{\parskip}{0pt}%
    }%
  {\end{enumerate}}
  
  \usepackage{CJKutf8}

\title{A Hierarchical Neural Autoencoder for Paragraphs and Documents}

\author{Jiwei Li, Minh-Thang Luong and Dan Jurafsky\\
{\normalsize Computer Science Department, Stanford University, Stanford, CA 94305, USA}\\
{\normalsize jiweil, lmthang, jurafsky@stanford.edu}
}

\begin{document}
\maketitle
\begin{abstract}
Natural language generation of coherent long texts like paragraphs or longer
documents is a challenging problem for recurrent networks models.
In this paper, we explore an important step toward  this generation task:
training an LSTM (Long-short term memory) auto-encoder to preserve and reconstruct multi-sentence paragraphs.
We introduce an LSTM model that hierarchically builds an embedding
for a paragraph from embeddings for sentences and words, then decodes
this embedding to reconstruct the original paragraph. We evaluate
the reconstructed paragraph using standard metrics like ROUGE and Entity Grid,
showing that neural models are able to encode texts in a way that preserve syntactic, semantic, and discourse coherence.
While only a first step toward generating coherent text units from neural models,
our work has the potential to significantly impact natural language generation and summarization\footnote{Code for models described in this paper are available at \url{www.stanford.edu/~jiweil/}.}.

\end{abstract}


\section{Introduction}
Generating coherent text is a central task in natural language processing.
A wide variety of theories exist for representing relationships between text units,
such as Rhetorical Structure Theory \cite{mann1988rhetorical} or
Discourse Representation Theory \cite{lascarides1991discourse},
for extracting these relations from text units
\cite[inter alia]{marcu2000rhetorical,lethanh2004generating,hernault2010hilda,feng2012text},
and for extracting other coherence properties characterizing
the role each text unit plays with others in a discourse
\cite[inter alia]{barzilay2008modeling,barzilay2004catching,elsner2008coreference,limodel}.
However, applying these to text generation remains difficult.
To understand how discourse units are connected,
one has to understand the communicative function of each unit, and
the role it plays within the context that encapsulates it, 
recursively all the way up for the entire text.
Identifying increasingly sophisticated
human-developed features may be insufficient for capturing 
these patterns. 
But developing neural-based alternatives has also been difficult.
Although neural 
representations for sentences can capture aspects of
coherent sentence structure 
\cite{ji2014representation,li2014recursive,limodel},
it's not clear how they could help in generating more broadly coherent text.

Recent  LSTM models \cite{hochreiter1997long} 
have shown powerful results on generating meaningful and grammatical sentences
in sequence generation tasks like machine translation \cite{sutskever2014sequence,bahdanau2014neural,luong2014addressing} or parsing \cite{vinyals2014grammar}. This performance
is at least partially attributable to the ability of these systems
to capture local compositionally: the way neighboring words are combined
semantically and syntactically to form meanings that they wish to express. 

Could these models be extended to deal with generation of larger structures
like paragraphs or even entire documents?
In standard sequence-to-sequence generation tasks,
an input sequence is mapped to a vector embedding that represents 
the sequence, and then to an output string of words.
Multi-text generation tasks like summarization 
could work in a similar way: the system reads a collection of input sentences, 
and is then asked to generate meaningful texts with certain properties 
(such as---for summarization---being succinct and conclusive).
Just as the local semantic and syntactic compositionally of words can be captured by LSTM models,
can the compositionally of discourse releations
of higher-level text units (e.g., clauses, sentences, paragraphs, and documents)
be  captured in a similar way, with clues about how text
units connect with each another stored in the neural compositional matrices?

In this paper we explore a first step toward this task of neural natural language generation.
We focus on the component task of
training a paragraph (document)-to-paragraph (document) autoencoder to
reconstruct the input text sequence from a compressed vector representation from a deep learning model. 
We develop hierarchical LSTM models that arranges tokens, sentences and paragraphs in a hierarchical structure,
with different levels of LSTMs  capturing compositionality at the
token-token and sentence-to-sentence levels. 

We offer in 
the following section to a brief description of sequence-to-sequence LSTM models. 
The proposed hierarchical LSTM models are then described in Section 3,
followed by experimental results in Section 4, and then
a brief conclusion.

\section{Long-Short Term Memory (LSTM)}

In this section we give a quick overview of LSTM models.
LSTM models \cite{hochreiter1997long} are defined as follows: 
given a sequence of inputs  $X=\{x_1,x_2,...,x_{n_X}\}$, an LSTM associates each timestep with an input, memory and output gate, 
respectively denoted as $i_t$, $f_t$ and $o_t$.
For notations, we disambiguate $e$ and $h$ where $e_t$ denote the vector for individual text unite (e.g., word or sentence) at time step t while $h_t$ denotes the vector computed by LSTM model at time t by combining $e_t$ and $h_{t-1}$. 
$\sigma$ denotes the sigmoid function. The vector representation $h_t$ for each time-step $t$ is given by:

\begin{equation}
\Bigg[
\begin{array}{lr}
i_t\\
f_t\\
o_t\\
l_t\\
\end{array}
\Bigg]=
\Bigg[
\begin{array}{c}
\sigma\\
\sigma\\
\sigma\\
\text{tanh}\\
\end{array}
\Bigg]
W\cdot
\Bigg[
\begin{array}{c}
h_{t-1}\\
e_{t}\\
\end{array}
\Bigg]
\end{equation}
\begin{equation}
c_t=f_t\cdot c_{t-1}+i_t\cdot l_t\\
\end{equation}
\begin{equation}
h_{t}^s=o_t\cdot c_t
\end{equation}
where $W\in \mathbb{R}^{4K\times 2K}$
In sequence-to-sequence generation tasks, each input $X$ is paired with a sequence of outputs to predict: 
$Y=\{y_1,y_2,...,y_{n_Y}\}$. 
An LSTM defines a distribution over outputs 
and sequentially predicts tokens using a softmax function:
\begin{equation}
\begin{aligned}
&P(Y|X)\\
&=\prod_{t\in [1,n_y]}p(y_t|x_1,x_2,...,x_t,y_1,y_2,...,y_{t-1})\\
&=\prod_{t\in [1,n_y]}\frac{\exp(f(h_{t-1},e_{y_t}))}{\sum_{y'}\exp(f(h_{t-1},e_{y'}))}
\end{aligned}
\label{equ-lstm}
\end{equation}
$f(h_{t-1}, e_{y_t})$ denotes the activation function between $e_{h-1}$ and $e_{y_t}$, where $h_{t-1}$ is the representation outputted from the LSTM at time $t-1$. 
Note that 
each sentence ends up with a special end-of-sentence symbol  $<$end$>$. 
Commonly, the input and output use two different LSTMs with different 
sets of convolutional parameters for capturing different compositional patterns. 

In the decoding procedure, the algorithm terminates when an $<$end$>$ token is predicted.
At each timestep, either a greedy approach or beam search can be adopted for word prediction.
Greedy search
selects the token with the largest conditional probability,
the embedding of which is then combined with preceding output for next step token prediction. 
For beam search, \cite{sutskever2014sequence} discovered that a beam size of 2 suffices to provide most of benefits of beam search.

\section{Paragraph Autoencoder}
In this section, we introduce our proposed hierarchical LSTM model for the autoencoder.

\subsection{Notation}
Let $D$ denote a paragraph or a document, which is comprised of 
a sequence of $N_D$ sentences, $D=\{s^1,s^2,...,s^{N_D}, {end}_D\}$. 
An additional ''$end_D$" token is appended to each document.
Each sentence $s$ is comprised of a sequence of tokens $s=\{w^1,w^2,...,w^{N_s}\}$ where $N_s$ denotes the length of the sentence, each sentence ending with an ``$end_s$" token.
The word $w$ is associated with a $K$-dimensional embedding $e_w$, $e_w=\{e_w^1, e_w^2, ..., e_w^{K}\}$.
Let $V$ denote vocabulary size. 
Each sentence $s$ is associated with a K-dimensional representation $e_s$. 

An autoencoder is a neural model where output units are directly connected with or identical to input units.
Typically, inputs are  compressed into a representation using neural models (encoding),  which is then used to reconstruct it back (decoding). 
For a paragraph autoencoder, both the input $X$ and output $Y$ are the same document $D$.
The autoencoder first compresses $D$ into a vector representation $e_D$ and then reconstructs $D$ based on $e_D$.

For simplicity, we define
$LSTM(h_{t-1},e_t)$ to be the LSTM operation on vectors $h_{t-1}$ and $e_t$ to achieve $h_t$ as in Equ.1 and 2.
For clarification, we first describe the following notations used in encoder and decoder:
\begin{itemize}
\item $h_t^w$ and $h_t^s$ denote hidden vectors from LSTM models, the subscripts of which indicate timestep $t$, the superscripts of which indicate operations at 
word level (w) or sequence level (s). $h_t^s(\text{enc})$ specifies encoding stage and $h_t^s(\text{dec})$ specifies decoding stage. 
\item $e_t^w$ and $e_t^s$ denotes word-level and sentence-level embedding for
word and
 sentence at position $t$ in terms of its residing sentence  or document.
\end{itemize}

\begin{figure*} [!ht]
\centering
\includegraphics[width=4.5in]{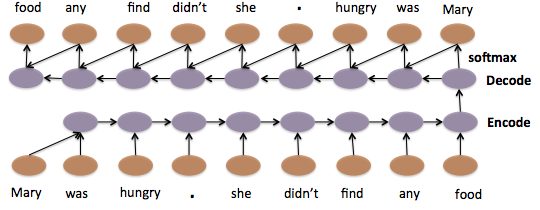}
\caption{Standard Sequence to Sequence Model.}\label{standard}
\end{figure*}
\begin{figure*} [!ht]
\centering
\includegraphics[width=4.5in]{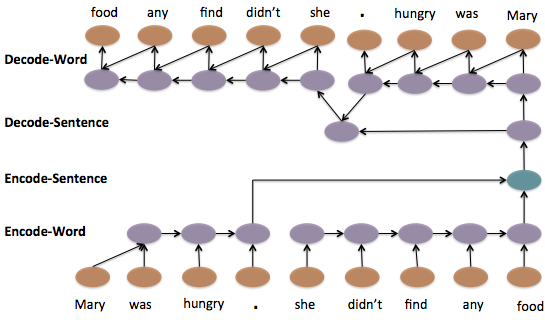}
\caption{Hierarchical Sequence to Sequence Model.}\label{hierarchical}
\end{figure*}
\begin{figure*} [!ht]
\centering
\includegraphics[width=4.5in]{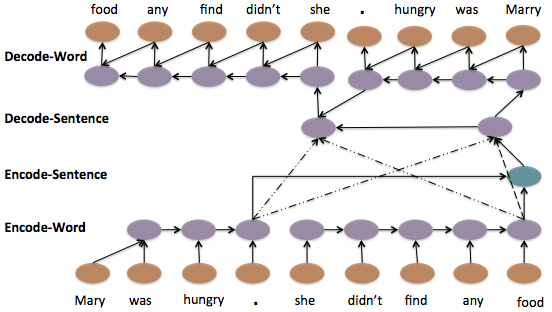}
\caption{Hierarchical Sequence to Sequence Model with Attention.}\label{bier-attention}
\end{figure*}

\subsection{Model 1: Standard LSTM}
The whole input and output are treated as one sequence of tokens. 
Following \newcite{sutskever2014sequence} and \newcite{bahdanau2014neural}, 
we trained an autoencoder that first maps input documents into vector representations
from a $LSTM_{\text{encode}}$ 
 and then 
reconstructs inputs by 
 predicting tokens within the document sequentially from a  $LSTM_{\text{decode}}$.
Two separate LSTMs are implemented for encoding and decoding with no sentence structures considered. 
Illustration is shown in Figure \ref{standard}.
\subsection{Model 2: Hierarchical LSTM}
The hierarchical model draws on
the intuition that just as the juxtaposition of words creates a joint meaning of a sentence,
the juxtaposition of sentences also creates a joint meaning of a paragraph or a document.

\paragraph{Encoder}
We first obtain representation vectors at the sentence level by
putting  one layer of LSTM (denoted as  $LSTM_{\text{encode}}^{\text{word}}$) on top of its containing words:
\begin{equation}
\begin{aligned}
&h_t^w(\text{enc})=LSTM_{\text{encode}}^{\text{word}}(e_{t}^w,h_{t-1}^w(\text{enc}))\\
\end{aligned}
\end{equation}
The vector output at the ending time-step is used to represent the entire sentence as 
$$e_s=h_{{end_s}}^w$$

To build representation $e_D$ for the current document/paragraph $D$, 
another layer of LSTM (denoted as $LSTM_{\text{encode}}^{\text{sentence}}$) 
is placed on top of all sentences, computing representations sequentially for each timestep:
\begin{equation}
\begin{aligned}
&h_t^s(\text{enc})=LSTM_{\text{encode}}^{\text{sentence}}(e_t^s,h_{t-1}^s(\text{enc}))\\
\end{aligned}
\end{equation}
Representation $e_{{end_D}}^s$ computed at the final time step is used to represent the entire document:
$e_D=h_{{end_D}}^s$. 

Thus one LSTM operates at the token level, leading to the acquisition of sentence-level representations that
are then used as inputs into the second LSTM that acquires document-level representations, 
in a hierarchical structure.

\paragraph{Decoder}
As with encoding, the decoding algorithm  operates on a hierarchical structure with two layers of LSTMs.
LSTM outputs at sentence level for time step $t$ are obtained by:
\begin{equation}
\begin{aligned}
&h_{t}^s(\text{dec})=LSTM_{\text{decode}}^{\text{sentence}}(e_{t}^s,h_{t-1}^s(\text{dec}))\\
\end{aligned}
\end{equation}
The initial time step $h_0^s(d)=e_D$, the end-to-end output from the encoding procedure. 
$h_{t}^s(d)$ is  used as the original input
into $LSTM_{\text{decode}}^{word}$ for subsequently 
predicting 
tokens within sentence $t+1$. 
$LSTM_{\text{decode}}^{word}$  
predicts tokens at each position sequentially,
the embedding of which is then combined with earlier hidden vectors for 
the next time-step prediction until the $end_s$ token is predicted. 
The procedure can be summarized as follows:
\begin{equation}
h_{t}^w(\text{dec})=LSTM_{\text{decode}}^{\text{sentence}}(e_{t}^w,h_{t-1}^w(\text{dec}))
\end{equation}
\begin{equation}
p(w|\cdot)=\text{softmax}(e_{w}, h_{t-1}^w(\text{dec}))
\end{equation}

During decoding, $LSTM_{\text{decode}}^{word}$ generates each word token
$w$ sequentially and combines it with earlier LSTM-outputted hidden
vectors.   The LSTM hidden vector computed at the final time step is used
to represent the current sentence.

This is passed to $LSTM_{\text{decode}}^{sentence}$, combined with $h_{t}^s$ for the acquisition of $h_{t+1}$, and outputted to the next time step in sentence decoding. 

For each timestep $t$, 
$LSTM_{\text{decode}}^{sentence}$ 
has to first decide 
whether decoding should proceed or come to a full stop: 
we add an additional token $\text{end}_D$ to the vocabulary. Decoding terminates when token $\text{end}_D$ is predicted. 
Details are shown in Figure \ref{hierarchical}.  

\subsection{Model 3: Hierarchical LSTM with Attention}
Attention models adopt a look-back strategy by linking the current decoding stage with 
input sentences in an attempt to consider which part of the input is most responsible for 
the current decoding state.
This attention version of hierarchical model is inspired by similar work in image caption generation and machine translation \cite{xu2015show,bahdanau2014neural}. 

Let $H=\{h_1^s(e), h_2^s(e), ..., h^s_{N}(e)\}$ be the collection of sentence-level hidden vectors for each sentence from the inputs, outputted from $LSTM_{\text{encode}}^{\text{Sentence}}$. Each element in H 
contains information about  input sequences
with a strong focus on the parts surrounding each specific sentence (time-step).
During decoding, suppose that
$e_{t}^s$ denotes the sentence-level embedding at current step and that
 $h_{t-1}^s(\text{dec})$ denotes the hidden vector outputted from $LSTM_{decode}^{sentence}$
at previous time step $t-1$. 
Attention models would first link the current-step decoding information, i.e., $h_{t-1}^s(\text{dec})$ which is outputted from  $LSTM_{dec}^{sentence}$ with each of the input sentences $i\in [1, N]$, characterized by a strength indicator $v_i$:
\begin{equation}
v_i=U^T f(W_1\cdot h_{t-1}^s(\text{dec})+W_2\cdot h_i^s(\text{enc}))
\end{equation}
$W_1, W_2\in \mathbb{R}^{K\times K}$, $U\in \mathbb{R}^{K\times 1}$. 
$v_i$ is then normalized:
\begin{equation}
\begin{aligned}
a_i=\frac{\exp(v_i)}{\sum_{i'}\exp(v_i')}\\
\end{aligned}
\end{equation}
The attention vector  is then created by averaging weights over all input sentences: 
\begin{equation}
m_t=\sum_{i\in[1,N_D]}a_i h_i^s(\text{enc})
\end{equation}
LSTM hidden vectors for current step is then achieved by combining $c_t$, $e_{t}^s$ and  $h_{t-1}^s(\text{dec})$:
\begin{equation}
\Bigg[
\begin{array}{lr}
i_t\\
f_t\\
o_t\\
l_t\\
\end{array}
\Bigg]=
\Bigg[
\begin{array}{c}
\sigma\\
\sigma\\
\sigma\\
\text{tanh}\\
\end{array}
\Bigg]
W\cdot
\Bigg[
\begin{array}{c}
h_{t-1}^s(\text{dec})\\
e_{t}^s\\
m_t\\
\end{array}
\Bigg]
\end{equation}
\begin{equation}
c_t=f_t\cdot c_{t-1}+i_t\cdot l_t\\
\end{equation}
\begin{equation}
h_{t}^s=o_t\cdot c_t
\end{equation}
where $W\in \mathbb{R}^{4K\times 3K}$. $h_t$ is
then used for word predicting as in the vanilla version of the hierarchical model. 
 
\subsection{Training and Testing}
Parameters are estimated  by maximizing likelihood of outputs given inputs, similar to standard sequence-to-sequence models. 
A softmax function is adopted for predicting each token within output documents, the error of which is first back-propagated through $LSTM_{\text{decode}}^{word}$ to sentences, then through $LSTM_{\text{decode}}^{sentence}$ to document representation $e_D$,
and last through
 $LSTM_{\text{encode}}^{sentence}$ and $LSTM_{\text{encode}}^{word}$ to 
inputs. 
Stochastic
gradient descent with minibatches 
is adopted. 

For testing, we adopt a greedy strategy with no beam search. 
For a given document $D$, $e_D$ is first obtained given already learned LSTM$_{\text{encode}}$ parameters and word embeddings. Then in decoding, $LSTM_{\text{decode}}^{\text{sentence}}$ computes embeddings at each sentence-level time-step,
which is first fed into the binary classifier to decide whether sentence decoding terminates and 
then into $LSTM_{\text{decode}}^{\text{word}}$ for word decoding. 

\section{Experiments}
\subsection{Dataset}
We implement the proposed autoencoder on two datasets, a highly domain specific dataset consisting of hotel reviews and a general dataset extracted from Wkipedia.

\paragraph{Hotel Reviews}  We use a subset of hotel reviews crawled
from TripAdvisor. We consider only reviews consisting sentences ranging from 50 to 250 words;
the model has problems dealing with extremely long sentences, as we will discuss later.
We keep a vocabulary set consisting of the 25,000 most frequent words.
A special ``$<$unk$>$" token is used to denote all the remaining less frequent
tokens.  Reviews that consist of more than 2 percent of unknown words are
discarded.  Our training dataset is comprised of roughly 340,000
reviews; the testing set is comprised of 40,000 reviews. Dataset
details are shown in Table \ref{dataset}.
\paragraph{Wikipedia}
We extracted paragraphs from Wikipedia corpus that meet the aforementioned length requirements.
We keep a top frequent vocabulary list of 120,000 words. 
Paragraphs with larger than 4 percent of unknown words are discarded. 
The training dataset is comprised of roughly 500,000 paragraphs and testing contains roughly 50,000. 

\begin{table}
\centering
\begin{tabular}{cccc}\hline
dataset&S per D&W per D&W per S\\\hline
Hotel-Review&8.8&124.8&14.1\\\hline
Wikipedia&8.4 &132.9&14.8 \\\hline
\end{tabular}
\caption{ Statistics for the Datasets. W, S and D respectively represent number of words, number of sentences, and number of documents/paragraphs.  For example, ``S per D" denotes average number of sentences per document.}
\label{dataset}
\end{table}

\subsection{Training Details and Implementation}

Previous research has shown that deep LSTMs work better than shallow ones for sequence-to-sequence
tasks \cite{vinyals2014grammar,sutskever2014sequence}.
We adopt a LSTM structure with four layer for encoding and four layer for decoding, each of which is comprised of a different set of parameters. 
Each LSTM layer consists of 1,000 hidden neurons and the dimensionality of word embeddings is set to 1,000. 
Other training details are given below, some of which follow \newcite{sutskever2014sequence}. 
\begin{tightitemize}
\item Input documents are reversed. 
\item LSTM parameters and word embeddings are initialized from a uniform distribution between [-0.08, 0.08].
\item Stochastic gradient decent is implemented without momentum using a fixed learning rate of 0.1. 
We stated halving the learning rate every half epoch after 5 epochs. We trained our models
for a total of 7 epochs.
\item Batch size is set to 32 (32 documents). 
\item Decoding algorithm allows generating at most 1.5 times the number of words in inputs.  
\item 0.2 dropout rate.
\item Gradient clipping is adopted by  scaling gradients when the norm exceeded a threshold of 5. 
\end{tightitemize}
Our implementation on a single GPU\footnote{Tesla K40m, 1 Kepler GK110B, 2880 Cuda cores.} processes a speed of approximately 600-1,200 tokens per second. 
We trained our models
for a total of 7 iterations.   
\subsection{Evaluations}
We need to measure the closeness of the output (candidate) to the input (reference). 
We first adopt two standard evaluation metrics, ROUGE \cite{lin2004rouge,lin2003automatic} and BLEU \cite{papineni2002bleu}. 
\paragraph{ROUGE}
is a recall-oriented  measure widely used in the summarization literature. 
It measures the n-gram recall between the
candidate text and the reference text(s).
In this work, we only have one reference document (the input document) and ROUGE score is therefore given by:
\begin{equation}
\text{ROUGE}_n=\frac{\sum_{\text{gram}_n\in \text{input}}\text{count}_{\text{match}}(\text{gram}_n)}{\sum_{\text{gram}_n\in \text{input}}\text{count}(\text{gram}_n)}
\end{equation}
where $\text{count}_{\text{match}}$
denotes the number of n-grams co-occurring in the input 
and output. 
We report ROUGE-1, 2 and W (based on weighted longest common
subsequence).
\paragraph{BLEU}Purely measuring recall will inappropriately  
reward long outputs. BLEU is designed to address such an issue by emphasizing precision. n-gram precision scores for our situation are given by:
\begin{equation}
\text{precision}_n=\frac{\sum_{\text{gram}_n\in \text{output}}\text{count}_{\text{match}}(\text{gram}_n) }{\sum_{\text{gram}_n\in \text{output}}\text{count}(\text{gram}_n)}
\end{equation}
BLEU then combines the average logarithm of precision scores with exceeded length penalization.
For details, see \newcite{papineni2002bleu}.

\begin{table*}[!ht]
\small
\centering
\begin{tabular}{|c|p{12cm}|}\hline
Input-Wiki&washington was unanimously elected President by the electors in both the 1788 – 1789 and 1792 elections . he oversaw the creation of a strong, well-financed national government that maintained neutrality in the french revolutionary wars , suppressed the whiskey rebellion , and won acceptance among Americans of all types . washington established many forms in government still used today , such as the cabinet system and inaugural address . his retirement after two terms and the peaceful transition from his presidency to that of john adams established a tradition that continued up until franklin d . roosevelt was elected to a third term . washington has been widely hailed as the " father of his country " even during his lifetime.\\\hline
Output-Wiki&
washington was elected as president in 1792 and  voters $<$unk$>$ of these two elections until 1789 . he continued suppression $<$unk$>$ whiskey rebellion of the french revolution war government , strong , national well are involved in the establishment of the fin advanced operations , won acceptance . as in the  government , such as the establishment of various forms of inauguration speech washington , and are still in use .  $<$unk$>$ continued after the two terms of his quiet transition to retirement of $<$unk$>$ $<$unk$>$ of tradition to have been elected to the third paragraph . but , " the united nations of the father " and in washington in his life , has been widely praised .\\\hline\hline
Input-Wiki&
apple inc . is an american multinational corporation headquartered in cupertino , california , that designs , develops , and sells consumer electronics , computer software , online services , and personal com - puters . its bestknown hardware products are the mac line of computers , the ipod media player , the iphone smartphone , and the ipad tablet computer . its online services include icloud , the itunes store , and the app store . apple's consumer software includes the os x and ios operating systems , the itunes media browser , the safari web browser , and the ilife and iwork creativity and productivity suites .
\\\hline
Output-Wiki&
apple is a us company in california , $<$unk$>$ , to develop electronics , softwares , and pc , sells . hardware include the mac series of computers , ipod , iphone . its online services , including icloud , itunes store and in app store . softwares , including os x and ios operating system , itunes , web browser , $<$ unk$>$ , including a productivity suite . \\\hline
Input-Wiki&
paris is the capital and most populous city of france . situated on the seine river , in the north of the country , it is in the centre of the le-de-france region . the city of paris has a population of 2273305 inhabitants . this makes it the fifth largest city in the european union measured by the population within the city limits . \\\hline
Output-Wiki&
paris is the capital and most populated city in france . located in the $<$unk$>$ , in the north of the country , it is the center of $<$unk$>$ . paris , the city has a population of $<$num$>$ inhabitants . this makes the eu ' s population within the city limits of the fifth largest city in the measurement .
\\\hline\hline
Input-Review&
on every visit to nyc , the hotel beacon is the place we love to stay . so conveniently located to central park , lincoln center and great local restaurants . the rooms are lovely . beds so comfortable , a great little kitchen and new wizz bang coffee maker . the staff are so accommodating and just love walking across the street to the fairway supermarket with every imaginable goodies to eat .
\\\hline
Output-Review&
every time in new york , lighthouse hotel is our favorite place to stay . very convenient , central park , lincoln center , and great restaurants . the room is wonderful , very comfortable bed , a kitchenette and a large explosion of coffee maker . the staff is so inclusive , just across the street to walk to the supermarket channel love with all kinds of what to eat .
\\\hline\hline
\end{tabular}
\caption{A few examples produced by the hierarchical LSTM alongside the inputs.}
\label{example}
\end{table*}

\paragraph{Coherence Evaluation} Neither BLEU nor ROUGE attempts to evaluate true
coherence. There is no generally  accepted and readily available coherence evaluation metric.\footnote{ 
\newcite{wolf2005representing} and \newcite{lin2011automatically} 
proposed metrics based on discourse relations, but these are hard to apply widely since
identifying discourse relations is a difficult problem.
Indeed sophisticated coherence evaluation metrics are seldom adopted in real-world applications,
and summarization 
researchers tend to use simple approximations like
number of overlapped tokens or  topic distribution similarity (e.g., \cite{yan2011evolutionary,yan2011timeline,celikyilmaz2011discovery}).}
Because of the difficulty of developing a
universal coherence evaluation metric, we proposed here only a 
tailored metric specific to our case. 
Based on the assumption that human-generated texts (i.e., input documents in our tasks) are coherent \cite{barzilay2008modeling},
we compare generated outputs with  input documents
in terms of how much original text order is preserved. 

We develop a grid evaluation metric similar to the entity transition algorithms in \cite{barzilay2004catching,lapata2005automatic}. 
The key idea of Barzilay and Lapata's models is to first identify grammatical roles (i.e., object and subject) that entities play and then model the transition probability over entities and roles across sentences.
We represent each sentence as a feature-vector consisting of verbs and nouns in the sentence. 
Next we align sentences from output documents to input sentences 
based on
sentence-to-sentence 
 F1 scores (precision and recall are computed similarly to ROUGE and BLEU but at sentence level) using feature vectors.
Note that multiple output sentences can be matched to one input sentence. 
 Assume that sentence $s_{\text{output}}^i$ is aligned with sentence $s_{\text{input}}^{i'}$, where $i$ and $i'$ denote position index for a output sentence  
and its aligned input. The penalization score $L$ is then given by:
\begin{equation}
\begin{aligned}
&L=\frac{2}{N_{\text{output}}\cdot (N_{\text{output}  }-1)} \\
&\times\sum_{  i\in [1,N_{\text{output}}-1]}\sum_{j\in [i+1,N_{\text{output}}]}  |(j-i)-(j'-i')|
\end{aligned}
\label{metric}
\end{equation} 
\begin{table*}
\centering
\begin{tabular}{cccccc}\hline
Model&Dataset& BLEU&ROUGE-1&ROUGE-2&Coherence(L)\\\hline
Standard&Hotel Review&0.241&0.571&0.302&1.92\\
Hierarchical&Hotel Review&0.267&0.590&0.330&1.71\\
Hierarchical+Attention&Hotel Review&0.285&0.624&0.355&1.57\\\hline\hline
Standard&Wikipedia&0.178&0.502&0.228&2.75\\
Hierarchical&Wikipedia&0.202&0.529&0.250&2.30\\
Hierarchical+Attention&Wikipedia&0.220&0.544&0.291&2.04\\\hline\hline
\end{tabular}
\caption{Results for three models on two datasets. As with coherence score L, smaller values signifies better performances.}
\label{result}
\end{table*}

Equ. \ref{metric} can be interpreted as follows: $(j-i)$ denotes the distance in terms of position index
between two outputted sentences indexed by $j$ and $i$,
and $(j'-i')$ denotes the distance between their mirrors in inputs.
As we wish to penalize the degree of permutation in terms of text order, we penalize the absolute difference between the two computed distances. 
This metric is also relevant to the overall performance of prediction and recall: an irrelevant output will be aligned
to a random input, thus 
being heavily penalized. The deficiency of the proposed metric is that it 
concerns itself only with a semantic perspective on coherence, barely
considering syntactical issues.

\subsection{Results}
A summary of our experimental results is given in Table \ref{result}. 
We observe better performances for the hotel-review dataset than the open domain Wikipedia dataset,
for the intuitive reason that
documents and sentences are written in a  more fixed format and easy to predict for hotel reviews.

The hierarchical model
that considers sentence-level structure
outperforms standard sequence-to-sequence models. Attention models
at the sentence level
introduce performance boost over vanilla hierarchical models. 


With respect to the coherence evaluation, 
the original sentence order is mostly preserved:
 the hierarchical model with attention achieves $L=1.57$ on the hotel-review dataset, equivalent to the fact that
the relative position of two input sentences are permuted by an average degree of 1.57. 
Even for the Wikipedia dataset where more poor-quality sentences are observed,
the original text order can still be adequately maintained with $L=2.04$. 

\section{Discussion and Future Work}
In this paper, we extended recent sequence-to-sequence LSTM models to the
task of multi-sentence generation.
We trained an autoencoder to see how well LSTM models can reconstruct input documents
of many sentences. We find that the proposed hierarchical LSTM models can partially
preserve the semantic and syntactic integrity of multi-text units and generate 
meaningful and grammatical sentences in coherent order.
Our model performs better than standard sequence-to-sequence models which do 
not consider the intrinsic hierarchical discourse structure  of texts.

While our work on auto-encoding for larger texts is only a preliminary
effort toward allowing neural models to deal with discourse,
it nonetheless suggests that 
neural models are capable of encoding 
complex clues about how coherent texts are connected .

The performance on this autoencoder task could certainly also benefit from 
more sophisticated neural models. For example one extension might
align the sentence currently being generated
with the original input sentence 
(similar to sequence-to-sequence translation in \cite{bahdanau2014neural}),
and later transform the original task to sentence-to-sentence generation.
However our long-term goal here is not on perfecting this
basic multi-text generation scenario of reconstructing input documents, 
but rather on extending it to more important applications.

That is, the autoencoder described in this work, where input sequence $X$ is identical to output $Y$,
is only the most basic instance of the family of document (paragraph)-to-document (paragraph) generation tasks. 
We hope the ideas proposed in this paper can play some role in enabling
such more sophisticated generation
tasks like summarization, where the inputs are original documents and outputs are summaries
or question answering, where inputs are questions and outputs are the actual wording of answers.
Sophisticated generation tasks like summarization or dialogue systems could 
extend this paradigm, and could themselves benefit from task-specific adaptations. 
In summarization, sentences to generate at each timestep might be
pre-pointed to or pre-aligned to specific aspects, topics, or pieces
of texts to be summarized. Dialogue systems could incorporate
information about the user or the time course of the dialogue.
In any case, we look forward to more sophi4d applications of
neural models to the important task of natural language generation.
\section{Acknowledgement}
The authors want to 
thank Gabor Angeli,  Sam Bowman, Percy Liang and other members of 
the Stanford NLP group for insightful comments and suggestion. 
We also thank
the three anonymous ACL reviewers for helpful
comments. 
This work is supported by Enlight Foundation Graduate Fellowship, and a gift from Bloomberg L.P, which we 
gratefully acknowledge.
\bibliographystyle{acl}
\bibliography{acl2013}

\end{document}